\documentclass[10pt,twocolumn,letterpaper]{article}

\usepackage{cvpr}      
 \usepackage[accsupp]{axessibility}
 
\usepackage{amsmath}
\usepackage{soul}
\usepackage{colortbl}
\usepackage{algorithm}
\usepackage{algorithmic}
\usepackage{amssymb}
\usepackage{mathtools}

\usepackage{booktabs}       
\usepackage{amsfonts}       
\usepackage{nicefrac}       
\usepackage{microtype}      
\usepackage{xcolor}         

\usepackage{caption}
\newtheorem{lemma}{Lemma}

\definecolor{cvprblue}{rgb}{0.21,0.49,0.74}
\usepackage[pagebackref,breaklinks,colorlinks,allcolors=cvprblue]{hyperref}

\def\paperID{17335} 
\def\confName{CVPR}
\def\confYear{2025}

\title{Preserving Clusters in Prompt Learning for Unsupervised Domain Adaptation}

\author{Tung-Long Vuong$^1$, Hoang Phan$^2$, Vy Vo$^1$, Anh Bui$^1$, Thanh-Toan Do$^1$, Trung Le$^1$, Dinh Phung$^1$\\\
$^1$Monash University,  $^2$New York University\\
{\tt\small \{Tung-Long.Vuong,v.vo,tuananh.bui,toan.do,trunglm,dinh.phung\}@monash.edu} \\ \tt\small hvp2011@nyu.edu}

\begin{document}
\maketitle
\begin{abstract}
Recent approaches leveraging multi-modal pre-trained models like CLIP for Unsupervised Domain Adaptation (UDA) have shown significant promise in bridging domain gaps and improving generalization by utilizing rich semantic knowledge and robust visual representations learned through extensive pre-training on diverse image-text datasets. While these methods achieve state-of-the-art performance across benchmarks, much of the improvement stems from base pseudo-labels (CLIP zero-shot predictions) and self-training mechanisms. Thus, the training mechanism exhibits a key limitation wherein the visual embedding distribution in target domains can deviate from the visual embedding distribution in the pre-trained model, leading to misguided signals from class descriptions. This work introduces a fresh solution to reinforce these pseudo-labels and facilitate target-prompt learning, by exploiting the geometry of visual and text embeddings - an aspect that is overlooked by existing methods. We first propose to directly leverage the reference predictions (from source prompts) based on the relationship between source and target visual embeddings. We later show that there is a strong clustering behavior observed between visual and text embeddings in pre-trained multi-modal models. Building on optimal transport theory, we transform this insight into a novel strategy to enforce the clustering property in text embeddings, further enhancing the alignment in the target domain. Our experiments and ablation studies validate the effectiveness of the proposed approach, demonstrating superior performance and improved quality of target prompts in terms of representation.




\end{abstract}    
\section{Introduction}
\label{sec:intro}

Significant advancements of deep learning have consistently enhanced performance on various tasks, albeit primarily within the confines of training data distributions. Therefore, deep models notoriously fail to maintain their performance when applied to new, unseen domains 
\cite{yosinski2014transferable, lones2021avoid, koh2021wilds, sagawa2019distributionally}. Unsupervised domain adaptation (UDA) offers a promising solution by utilizing labeled data from a source domain alongside unlabeled data from a target domain, thus circumventing the costly and labor-intensive process of annotating new domain data. In recent years, many domain adaptation methods have been proposed, typically focusing on learning invariant features across domains, as motivated by the theoretical analysis by \citet{ben2006analysis, ben2010theory}. These methods generally (i) minimize the discrepancy between feature representations
across domains \citep{long2015learning, long2017deep, sun2016deep, phan2023global} or (ii) adversarially
learn indistinguishable representations to fool a domain discriminator \citep{ganin2016domain, ganin2015unsupervised, tzeng2017adversarial,long2018transferable}. However,  focusing on aligning representations can inadvertently compromise model performance by diminishing discriminative features \citep{ge2023domain, tang2020unsupervised}. 

Recent developments have demonstrated the potential of leveraging pre-trained models such as CLIP \citep{radford2021learning} for the UDA problem, significantly bridging domain gaps and enhancing generalization. These benefits arise from the rich semantic and robust visual representations gained through extensive pre-training across diverse image-text datasets. Building on this approach, DAPL \cite{ge2023domain} innovatively introduces domain-specific and domain-agnostic prompts, enabling the efficient adaptation of pre-trained vision-language models without the need to fine-tune the entire model. Extending this training procedure to the multi-source domain adaptation scenario, MPA \citep{chen2022multiprompt} employs an auto-encoder to align training and target prompts while PGA \cite{phan2024enhancing} frames the learning from a multi-domain problem as a multi-objective optimization problem and align per-objective gradients to promote consensus across objectives. 

\paragraph{Contribution.} 
While prior vision-language UDA methods achieve state-of-the-art performance across benchmarks, much of the improvement stems from the use of base pseudo-labels (CLIP zero-shot predictions) and self-training mechanisms. However, they do have limitations. First, the visual embedding distribution in the target domain may differ from the textual embedding distribution in the pre-trained models, resulting in misaligned class descriptions when relying on the "over-generalized" nature of pre-trained models (such as CLIP zero-shot predictions) without incorporating domain-specific context. Secondly, previous methods primarily focus on aligning source-target prompts within the \textit{textual embedding space} to refine target prompts, placing less emphasis on the representation perspective, specifically the alignment between visual and text embeddings. Moreover, we observe that source-target prompt alignment has a limited impact on target prompts (Table~\ref{tab:pseudo-label}, Simple Alignment - Eq.~(\ref{eq:simple_baseline}): Even after alignment, the target prompt performs poorly. Further analysis is provided in Section~\ref{sec:ablation_study}.)

To address these challenges, we leverage reference predictions (from source prompts) based on the relationship between source and target visual embeddings. This helps enhance pseudo-labels and facilitates target-prompt learning. Additionally, drawing inspiration from the strong clustering behavior observed between visual and textual embeddings in pre-trained multi-modal models, we enforce clustering properties in text embeddings by minimizing the Wasserstein distance \cite{villani2008optimal} between the distribution of text prompts and the distribution of visual embeddings in the target domains. Our experiments and ablation studies validate the effectiveness of this approach, showing superior performance and improved quality of target prompts in terms of representation.


\section{Related work}

\textbf{Domain Adaptation (DA)} addresses the challenge of transferring knowledge learned from a source domain to a target domain, where both domains share the same label space but have different data distributions \cite{ben2010theory}. Deep domain adaptation has demonstrated impressive performance across various tasks through different approaches \citep{NEURIPS2018_717d8b3d, zhao2018multiple}. The fundamental concept is to minimize the distributional gap between source and target domains in a shared feature space by reducing distribution divergence, commonly measured through Jensen-Shannon divergence \cite{ganin2015unsupervised,TzengHDS15,shu2018dirt}, maximum mean discrepancy \cite{gretton2007kernel,long2015learning}, or Wasserstein distance \cite{shen2018wasserstein,SWD_CVPR19}.


\textbf{Unsupervised Domain Adaptation (UDA)} presents additional challenges as the target domain is unlabeled. Early UDA methods focused on learning domain-invariant features by minimizing domain discrepancy \cite{long2015learning,long2017deep,sun2016deep}. For instance, DAN \cite{long2015learning} aligns domains by minimizing maximum mean discrepancy on task-specific layers, while CORAL \cite{sun2016deep} aligns second-order statistics using linear projection. Inspired by GANs \cite{GAN_NIPS14}, adversarial learning emerged as another prominent approach. Methods like DANN \cite{ganin2015unsupervised} and CDAN \cite{long2017conditional} employ domain discriminators to distinguish source from target samples while training feature extractors to generate domain-invariant features. However, recent work has highlighted that while domain alignment can be successful, it may compromise class discrimination due to distorted semantic features \cite{cai2019learning,tang2020unsupervised}. Several methods have attempted to address this challenge through approaches like entropy-based regularization \cite{ETD_CVPR20} and batch nuclear-norm maximization \cite{BNM_CVPR2020}.

\textbf{Prompt Learning for Domain Adaptation.}
Prompt learning, initially introduced in NLP \cite{DBLP:conf/emnlp/PetroniRRLBWM19}, has recently been adapted for vision-language models \cite{radford2021learning,DBLP:conf/icml/JiaYXCPPLSLD21}. Rather than fine-tuning entire networks, prompt learning enables efficient adaptation to downstream tasks by tuning prompt parameters. Such notable models as CLIP \cite{radford2021learning} have demonstrated strong visual representations through pre-training on large-scale image-text pairs. In the context of UDA, prompt learning offers a novel direction to address domain adaptation challenges \cite{gal2022stylegan, zara2023autolabel, phancontrollable}. DAPL \cite{ge2023domain} leverages CLIP's generalization capabilities to learn both domain-agnostic and domain-specific prompts, effectively balancing domain alignment and classification performance through contrastive learning. Building on this idea, MPA \cite{chen2022multiprompt} extends prompt learning to multi-source UDA scenarios. MPA first learns from multiple source domains, then adapts the prompt learning strategy to each source-target domain pair. Prompts are aligned using a denoising auto-encoder with Euclidean distance. More recently, PGA \cite{phan2024enhancing} frames the multi-domain objective (including the target domain) as a multi-objective optimization problem, aligning per-objective gradients to promote consensus across objectives. PGA further incorporates gradient norm penalization to enhance prompt generalization.

\section{Background}
\label{sec:background}

\subsection{Unsupervised Domain Adaptation}
\label{sec:background0UDA}

Given a set of $N \geq 1$ source domains $\{D_{S_i}\}_{i=1}^N$ each of which is a collection of data-label pairs of domain $i$-th, i.e. $D_{S_i} = \{\boldsymbol{x}_j, y_j\}_{j=1}^{N_{S_i}}$, $y_j \in [K]:= \{1,...,K\}$, and one unlabelled target domain $D_T = \{\boldsymbol{x}_j\}_{j=1}^{N_T}$, where $N_{S_i}$ and $N_T$ are respectively the number of data points in source domain $D_{S_i}$ and target domain $D_T$, the goal is to learn a model that can perform well on the unlabelled target domain. In this paper, we focus on classification problems and denote $K$ as the number of classes. 



\subsection{CLIP-based Zero-Shot Classification.} 
\label{sec:background-zero-shot}

CLIP \citep{radford2021learning} is a vision-language model that consists of a visual encoder $f_v$ and a textual encoder $f_t$, which are trained to align the visual embedding $f_v(\boldsymbol{x})$ of an image $\boldsymbol{x}$ with the textual embedding $f_t(y)$ of the corresponding label $y$. The textual representation is derived from a manually crafted prompt $\boldsymbol{p}_k$ in the form ``A photo of a $[\text{CLASS}_k]$", where $[\text{CLASS}_k]$ is the $k$-th class's name.

In zero-shot inference, $K$ class names are forwarded through the text encoder $f_t$ and one with the highest similarity with the visual embedding $f_v(x)$ is the predicted class: 
\begin{equation}
    \hat{y} = \text{argmax}_{k}P(y|\boldsymbol{x},\boldsymbol{p}_k) 
    \label{zs_clip}
\end{equation}
where we have defined
\begin{align}
    P(y|\boldsymbol{x}, \boldsymbol{p}_k) = \frac{\exp(\langle f_v(\boldsymbol{x}), f_t(\boldsymbol{p}_k) \rangle/\gamma)}{\sum_{k'=1}^K\exp(\langle f_v(\boldsymbol{x}), f_t(\boldsymbol{p}_{k'})\rangle/\gamma)},
\end{align}
$\langle.,.\rangle$ measures cosine similarity and $\gamma$ is the temperature.

\subsection{Domain Adaptation via Prompt Learning}
\label{sec:background-prompt-learning}

\paragraph{Prompt Design.}
A common assumption in domain adaptation literature is that each domain can be represented by domain-specific features - which are found/presented only in this particular domain and domain-invariant features - which are shared with others \cite{chen2022multiprompt, phan2024enhancing}.

To reflect this, prompt-learning-based approaches as \cite{chen2022multiprompt, phan2024enhancing} utilize two sets of prompts for each domain $D_{S_i}$: \textit{domain-invariant prompt}  $\boldsymbol{P}_{sh}$ shared among all domains and \textit{domain-specific prompt} $\boldsymbol{P}_{S_i}$ specific to the domain $D_{S_i}$. Similarly, the target domain $D_{T}$ will be represented by $\boldsymbol{P}_{sh}$ and $\boldsymbol{P}_{T}$ prompts. 
More specifically, following \cite{chen2022multiprompt}, we do not just use single but multiple domain-invariant prompts $\{ \boldsymbol{P}_{sh}^k \}_{k=1}^K$, where each $\boldsymbol{P}_{sh}^k = [\boldsymbol{v}_1^k | \boldsymbol{v}_2^k | \cdots | \boldsymbol{v}_{M_1}^k]$ is a shared prompt among all domains and specific to class $k$-th. 
For domain-specific prompts $\boldsymbol{P}_{S_i}$ and $\boldsymbol{P}_{T}$, which are not shared among domains, are represented by $M_2$ tokens, i.e., $\boldsymbol{P}_{S_i} =  [\boldsymbol{u}_1^{S_i} | \boldsymbol{u}_2^{S_i} | \cdots | \boldsymbol{u}_{M_2}^{S_i}]$, $\boldsymbol{P}_T =  [\boldsymbol{u}_1^T | \boldsymbol{u}_2^T | \cdots | \boldsymbol{u}_{M_2}^T]$. 
Based on this, a prompt $p_k = [\boldsymbol{P}_{sh}^k][\boldsymbol{P}_{S_i}][\text{CLASS}_k]$ is to represent the predictive distribution of a sample of source domain $D_{S_i}$ belong to class $k$, and similarly $p_k = [\boldsymbol{P}_{sh}^k][\boldsymbol{P}_{T}][\text{CLASS}_k]$ for that of a target sample.

\paragraph{UDA via Prompt Learning with CLIP-Pseudo-Label.}
\label{sec:background-uda-prompt-learning}
To transfer knowledge from source domains to the target domain, \cite{chen2022multiprompt} proposed to fine-tune these prompts $\boldsymbol{P}_{sh}$, $\boldsymbol{P}_{S_i}, \boldsymbol{P}_{T}$, simultaneously, where $\boldsymbol{P}_{sh}$ captures the shared predictive features among domains while the prompts $\boldsymbol{P}_{S_i}, \boldsymbol{P}_{T}$ capture unique characteristics of each domain, by optimizing the objective: 
\begin{align}
        &\mathcal{L}_{total}(\boldsymbol{P}) = \sum_{i=1}^N \mathcal{L}_{S_i}(\boldsymbol{P}_{sh}, \boldsymbol{P}_{S_i}) +\mathcal{L}_T(\boldsymbol{P}_{sh}, \boldsymbol{P}_T) ,
        \label{eq:simple_baseline}
\end{align}
where objectives for source domains are defined as:
\begin{align}
     &\mathcal{L}_{S_i}(\boldsymbol{P}_{sh}, \boldsymbol{P}_{S_i}) = \text{CE}(\boldsymbol{P}_{sh}, \boldsymbol{P}_{S_i}; \boldsymbol{X}_{S_i}, Y_{S_i})\nonumber \\
     &= -\frac{1}{N_{S_i}}\sum_{j=1}^{N_{S_i}}\log P(y=y_j|\boldsymbol{x}_j, \boldsymbol{P}_{sh},\boldsymbol{P}_{S_i}), \text{and} \nonumber
\end{align}

For target domain, since labels are not available in the target domain, previous approaches,
such as DAPL \cite{ge2023domain}, MPA \cite{chen2022multiprompt}, and PGA \cite{phan2024enhancing}, use pre-trained CLIP models to generate pseudo-labels $\hat{y}$ as in Eq. (\ref{zs_clip}) for the target domain to train target classifiers and learn both domain-invariant and domain-specific prompts. Therefore, the objective for the target domain is as follows:

\begin{align}
    &\mathcal{L}_{T}(\boldsymbol{P}_{sh}, \boldsymbol{P}_{T}) = \text{CE}_{\alpha}(\boldsymbol{P}_{sh}, \boldsymbol{P}_{T}; \boldsymbol{X}_{T}, Y_{T}) \nonumber \\
        &= -\frac{1}{N_{T}}\sum_{j=1}^{N_{T}} \mathbb{I}_{(P(y=\hat{y}_j|\boldsymbol{x}_j) \geq \alpha)} \log P(y=\hat{y_j}|\boldsymbol{x}_j, \boldsymbol{P}_{sh},\boldsymbol{P}_{T})\nonumber    
\end{align}



The target objective is applied to target samples for which the zero-shot prediction confidence from CLIP exceeds a threshold $\alpha$. It is evident that when there is a significant covariate shift between the target domain and the dataset used to train the CLIP models, the accuracy of the pseudo-labels decreases markedly. For instance, as shown in Table~\ref{tab:pseudo-label}, the accuracy of CLIP’s zero-shot predictions varies significantly across domains in the OfficeHome dataset. This limitation forms one of the motivations for our approach, as presented in the following sections.  

\begin{table}[ht]
\centering
\resizebox{1.0\columnwidth}{!}{\begin{tabular}{@{}llcccc@{}}
\toprule
Setting & Inference & Ar & Cl & Pr & Rw \\
\midrule
\rowcolor[gray]{0.9} \multicolumn{6}{c}{Training on \textbf{source} data only} \\
\midrule 
\textbf{Zero-shot} from & CLIP \citep{radford2021learning}  &71.2 & 50.4 & 81.4 & 82.6 \\
\midrule
\textbf{Source-combined} & Source prompt &72.2 & 55.9 & 82.6 & 83.3 \\
\midrule

 & Average prompt &74.3 & 57.4 & 84.5 & 84.7 \\
\cmidrule{2-6}
&Ar prompt & - & 56.0 & 80.0 & 82.1 \\
\textbf{Multi-source} &Cl prompt &70.4 & - & 81.1 & 80.4 \\
&Pr prompt &70.2 & 55.0 & - & 84.1 \\
&Wr prompt &73.9 & 55.5 & 84.5 & - \\
\midrule
\rowcolor[gray]{0.9} \multicolumn{6}{c}{Supervised training on \textbf{target} domain only} \\
\midrule
 & Target prompt  &99.6 & 91.6 & 99.3 & 98.6 \\
 \midrule
\rowcolor[gray]{0.9} \multicolumn{6}{c}{Trainining both source-target domains as Eq.~(\ref{eq:simple_baseline})} \\
\midrule
 & Target prompt  &47.6 & 29.0 & 53.5  & 63.5 \\
  & Source prompt  &74.3 & 57.2 & 84.3 &84.8 \\
\bottomrule
\end{tabular}
}
\caption{Analyses of prompts performance on OfficeHome dataset. \textbf{Source-combined} setting, where data from all source domains are merged, and a \textbf{multi-source} setting, which utilizes individual domain identifications. \label{tab:pseudo-label}}
\end{table}

\begin{figure*}[h!]
\begin{centering}
\centering{}\includegraphics[width=.9\linewidth]{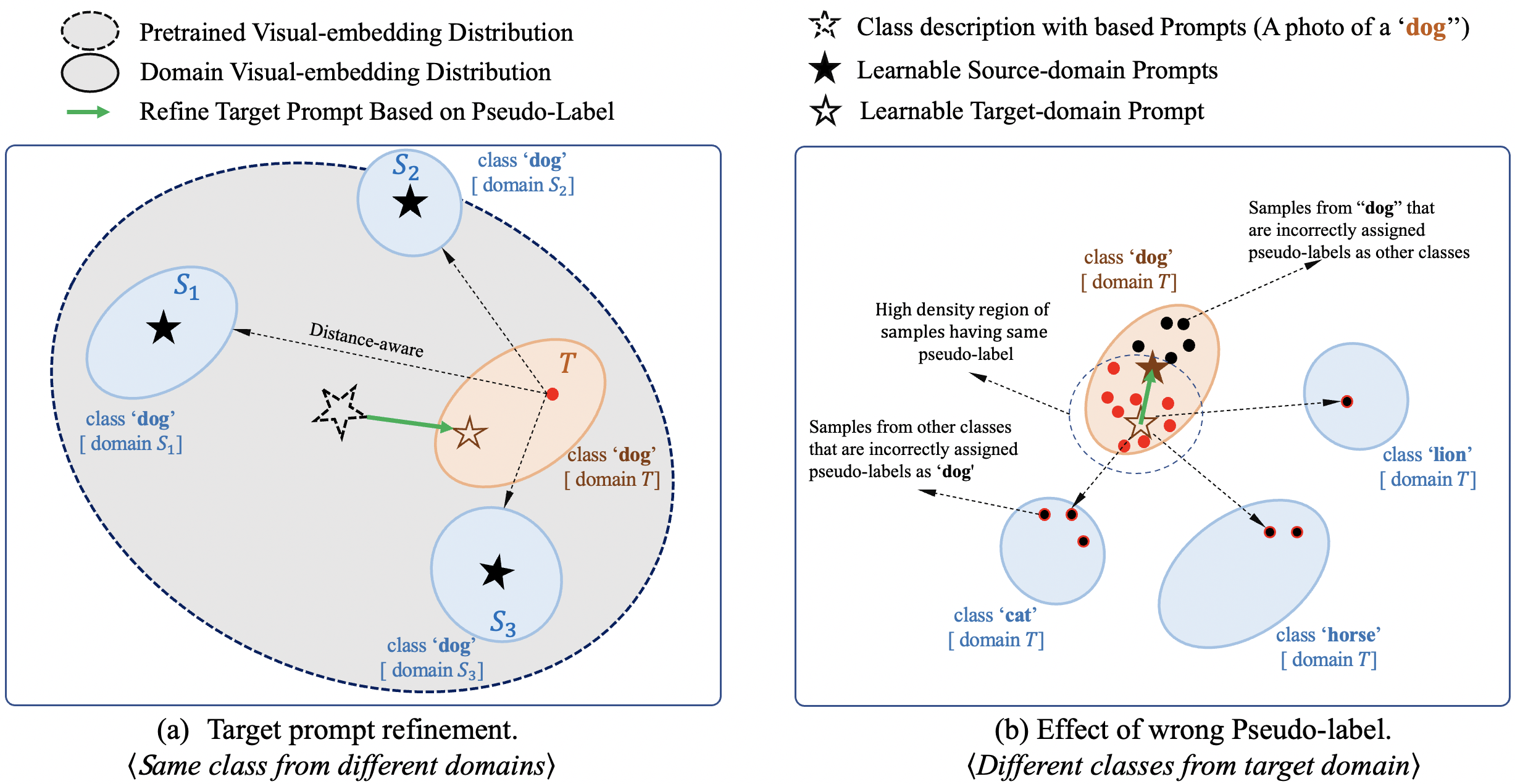}
\par\end{centering}
\caption{ (a) \textbf{Target-domain prompt refinement.} 
Pre-trained models are trained on large-scale datasets, which are assumed to fully or partially cover the data distribution of downstream tasks. For example, “\textit{dog} with style domain $S_{1,...,3,T}$” can be viewed as subdomains of “\textit{dog}”. However, the visual embedding distribution in training may shift from that of the pretraining distribution, causing class descriptions without domain-specific context to become misaligned. This shift necessitates alignment using domain-specific contexts. While the source domain can leverage labeled data to refine its domain-specific context, the target domain lacks this advantage. A common approach to address this limitation is to incorporate domain-specific contexts and leverage zero-shot predictions from the CLIP model to guide alignment between domain-specific contexts and visual-representation in the target domain.
(b) \textbf{Effect of incorrect pseudo-label.}  Behavior of prompt learning with incorrect pseudo-labels in a supervised manner on the target domain. 
The flat regions in gray and orange represent areas in the embedding space corresponding to different classes, each centered on its optimal domain-specific prompt embedding (with class “\textit{dog}” in orange and other classes in blue).
Red points indicate samples from class “\textit{dog}” with correct pseudo-labels, while black points represent samples from class “\textit{dog}” that have been incorrectly pseudo-labeled. Additionally, some samples from regions belonging to other classes (gray areas) are mislabeled as class “\textit{dog}”, shown as black points with red borders. 
The text embeddings learned from these pseudo-labels fail to align well with the true target embeddings; instead, they form a cluster around a single point—the centroid of all samples labeled as class “\textit{dog}” (both correctly and incorrectly labeled)—remaining distant from the actual target embeddings.}
\label{fig:space}
\end{figure*}

\section{Motivations}
\label{sec:motivation}

\paragraph{Limitations of existing pseudo-label guidance.} As described above, we design shared or domain-invariant prompts $\boldsymbol{P}_{sh}$ to capture embeddings that are consistent across both source and target domains, while we use domain-specific prompts $\boldsymbol{P}_{S_i}$ to capture unique characteristics of each source domain. The domain-invariant prompts $\boldsymbol{P}_{sh}$ enable learning of shared knowledge across source domains, which can then be transferred to the target domain. Additionally, to train both domain-invariant prompts $\boldsymbol{P}_{sh}$ and target-specific prompts $\boldsymbol{P}_T$ with a focus on the target domain, most recent state-of-the-art approaches~\cite{ge2023domain, phan2024enhancing} leverage pseudo-labels obtained through zero-shot inference using CLIP models. However, we argue that generating reliable and accurate pseudo-labels can be challenging when relying on zero-shot predictions from CLIP models, particularly \textit{if there is a significant covariate shift between the target domain and the dataset on which the CLIP models were trained}. Furthermore, pseudo-labels are crucial for guiding the target text embeddings $\boldsymbol{\tau}^k_T$ for each class $k$ by aligning them with the visual embeddings from the corresponding target class. This raises the question: \textit{How can we enhance these critical pseudo-labels to improve the target target prompt?} 

\paragraph{Enhancing pseudo-labels using source prompts.} When training the source and target prompt, we use domain-invariant prompts $\boldsymbol{P}_{sh}$ shared across both source and target domains to capture the domain-invariant embeddings common to data in both domains. As a result, the source prompts can encapsulate domain-invariant embeddings of the target data. Additionally, the source and target data share certain common embeddings, making it reasonable to utilize source prompts to generate pseudo-labels for the target domain. To illustrate the advantage of using aggregated source prompts over standalone CLIP models in generating more accurate pseudo-labels, we conduct experiments on the OfficeHome dataset. Table~\ref{tab:pseudo-label} (Training on source data only setting) presents the accuracies of the average aggregated source prompts (multi-source) and single source prompt (source combined) on the target domain, where only source prompts are used for predictions. The results indicate that aggregated source prompts produce more accurate pseudo-labels, hence further improving predictions. This observation aligns with our motivation and supports the proposed approach. 
Additionally, the transferability of different source domains to the target domain varies based on their similarity (Table~\ref{tab:pseudo-label}: Training on source data only: Multi-source). This observation motivates the design of a distance-aware transfer strategy to combine references from source prompts and base prompt for generating more accurate pseudo-labels, as detailed in Section~\ref{sec:prompt_enhance}.

Furthermore, we observe that while more accurate pseudo-labels help guide the text embeddings more effectively toward their class clusters, they still introduce inaccuracies. These inaccuracies cause the text embeddings to skew toward regions of correct pseudo-labels within their clusters, as illustrated in Figure \ref{fig:space}, ultimately compromising performance on entire clusters. Thus, it is essential to introduce an additional component to correct this bias and maintain consistent predictions across clusters.

\paragraph{Clustering preservation prediction.} According to the clustering assumption, data tend to be organized into clusters. To verify this assumption for visual embeddings from CLIP models, we assume we have labels for the target domain and train a target prompt using only class-specific target prompts (i.e., $[\boldsymbol{P}^k_T][CLASS_k]$) on the target domain. This setup implies that for each class $k$, we rely on a single text embedding, $\boldsymbol{\tau}^k_T$, to represent its visual embeddings. The high accuracies obtained, as shown in Table~\ref{tab:pseudo-label} (supervised training on target domain only), demonstrate that the visual embeddings for each class using CLIP models tend to form a single cluster in the embedding space.

Although using source prompts to generate pseudo-label guidance for the target prompt outperforms zero-shot predictions from CLIP models, incorrect pseudo-labels or the absence of pseudo-label guidance may still occur for lower-confidence predictions. Consequently, in the cluster for class $k$ in the target domain, there are visual embeddings with correct pseudo-labels and those with incorrect or missing pseudo-labels (shown in black in Figure~\ref{fig:space}b). As a result, the text embedding $\boldsymbol{\tau}^k_T$ for this class acts as a prototype or centroid for the red visual embeddings but may fail to represent the entire cluster for class $k$. This limitation could lead to incorrect predictions for the black visual embeddings and potential confusion with visual embeddings from other classes. To obtain text embeddings $\boldsymbol{\tau}^k_T$ that effectively represent the entire cluster for each class $k$ and ensure clustering-preserving predictions, we adopt the clustering viewpoint of optimal transport \cite{tuan_tidot, pmlr-v202-vuong23a}. Specifically, our objective is to minimize the optimal transport (Wasserstein-WS) distance between the distribution of visual embeddings in the target domain and the distribution of the text embeddings $\boldsymbol{\tau}^k_T$. As demonstrated in Lemma \ref{cor:distortion_data}, the text embeddings $\boldsymbol{\tau}^k_T$ act as prototypes or centroids of the clusters, leading to an alignment correction for the text embeddings, as shown in Figure 2. Moreover, minimizing the WS distance adjusts the text embeddings $\boldsymbol{\tau}^k_T$ to serve as accurate cluster centroids, while the pseudo-label guidance aids in aligning these embeddings with their correct class-$k$ clusters.    

\section{Proposed Method}

\textbf{Notation:} Since our method focuses on analyzing the representations; for simplicity, we denote $\boldsymbol{\tau}_{S_i}^k$ as the text embedding of the source domain-specific prompt, given by $\boldsymbol{\tau}_{S_i}^k=f_t\left([\boldsymbol{P}_{sh}^k][\boldsymbol{P}_{S_i}][\text{CLASS}_k]\right)$. Similarly,  $\boldsymbol{\tau}_{T}^k=f_t\left([\boldsymbol{P}_{sh}^k][\boldsymbol{P}_{T}][\text{CLASS}_k]\right)$ and $\boldsymbol{\tau}_{base}^k=f_t\left( \text{``A photo of a [} \text{CLASS}_k \text{]"}\right)$ as the text embeddings of the target prompt and base prompt, respectively.

\subsection{Prompt Learning with Source Domains}
Because labels are available in all source domains, we apply the cross-entropy loss to fine-tune these associated prompts $\boldsymbol{P}_{sh}, \boldsymbol{P}_{S_i}$ with the source-domain objective loss:

\begin{equation}
\mathcal{L}_{S} = \frac{1}{N}\sum_{i=1}^N\mathcal{L}_{S_i}    
\end{equation}
where we have defined
\begin{align}
    &\mathcal{L}_{S_i}(\boldsymbol{P}_{sh}, \boldsymbol{P}_{S_i}) = \text{CE}(\boldsymbol{P}_{sh}, \boldsymbol{P}_{S_i}; \boldsymbol{X}_{S_i}, Y_{S_i})\nonumber \\
     &= -\frac{1}{N_{S_i}}\sum_{j=1}^{N_{S_i}}\log P(y=y_j|\boldsymbol{x}_j, \boldsymbol{P}_{sh},\boldsymbol{P}_{S_i}) \label{src_loss}
\end{align}

\subsection{Prompt Learning with Target Domain}
\label{sec:prompt_enhance}
\textbf{Pseudo-label Enhancement.} Building on our motivation, we enhance the quality of pseudo-labels from zero-shot predictions on the target domain by incorporating predictions from the source domains. Specifically, we calculate the weighted average cosine distance from the visual embedding $\boldsymbol{z}=f_v(\boldsymbol{x})$ to the text embeddings of each class across all source domains. We then average this result with the text embedding from the base prompt as follows:

\begin{equation}
    \hat{y}[k] = \frac{\exp\left(\langle \boldsymbol{z}, \boldsymbol{\tau}_{ave}^k((x)) \rangle/\gamma\right )}{\sum_{k'=1}^K\exp\left(\langle \boldsymbol{z}, \boldsymbol{\tau}_{ave}^{k'}(x) \rangle/\gamma\right )}
    \label{eq:enhanced_label}
\end{equation}

where we have defined


\begin{equation}
   \boldsymbol{\tau}_{ave}^k(x) = \frac{1}{2}\boldsymbol{\tau}_{base}^k  + \frac{w_{k,i}(x)}{2}\sum_{i=1}^N \boldsymbol{\tau}_{S_i}^k
   \label{eq:enhance_prompt}
\end{equation}

and $w_{i,k}$ is the important weight for each domain-class. 

As discussed in Section~\ref{sec:motivation}, to compute \( w_{k,i}(x) \), we use the \( l_2 \) distance between the visual embedding \( z^{pre} \) (the unnormalized visual embedding of \( x \), where \( z = \frac{z^{pre}}{\| z^{pre} \|_2} \)) and the centroid of class \( k \) in domain \( i \). Let \( C^i = \{c^i_k\}_{k=1}^K \) represent the set of class centroids for the \( i \)-th domain i.e., $c^i_k = \frac{1}{\sum_{j=1}^{N_{S_i}}\mathbb{I}_{(y^i_j=k)}}\sum_{j=1}^{N_{S_i}}\mathbb{I}_{(y^i_j=k)}  \boldsymbol{z}^{pre,i}_j \nonumber$. The weight for class \( k \) is then calculated by applying the softmax function:

\begin{equation}
    w_{i,k}(z) = \frac{\exp\left ( \left \| z_{pre}-c^i_k \right \|_2\right )}{\sum_{i'=1}^N\exp\left(\left \| z-c^{i'}_k \right \|_2\right )}
\end{equation}
Note that the weights \(w_{i,k}(z)\) are applied to compute the combined prompts \(\boldsymbol{\tau}_{\text{ave}}^k\), rather than being applied directly to the outputs of the softmax (i.e., the predictions from individual prompts). This design preserves semantic similarity, as the magnitude of cosine similarity between the visual embedding and the text embedding of each individual prompt is taken into account in the final pseudo-label. Meanwhile, the weights \(w_{k,i}(x)\) represent the spatial relationship between the visual embedding and the text embedding.

\paragraph{Training target prompts with Pseudo-label Guidance.} Given data sample $x_j$ and its corresponding enhanced pseudo-label $\hat{y}_j$,  we minimize the following objective function to train the target prompts:
\begin{align}
    &\mathcal{L}_{T}(\boldsymbol{P}_{sh}, \boldsymbol{P}_{T}) = \text{CE}_{\tau}(\boldsymbol{P}_{sh}, \boldsymbol{P}_{T}; \boldsymbol{X}_{T}, Y_{T}) \nonumber \\
    &= -\frac{1}{N_{T}}\sum_{j=1}^{N_{T}} \sum_{k=1}^{K}\hat{y_j}[k]\log P(y=k|\boldsymbol{x}_j, \boldsymbol{P}_{sh},\boldsymbol{P}_{T})\label{tgt_loss} 
\end{align}
\subsection{Clustering Refined by Optimal Transport}

We propose minimizing the Wasserstein distance between the target prompts' text embeddings and the visual embeddings from the target domain. Specifically, denote \( \mathcal{T} = \{\boldsymbol{\tau}_{T}^k\}_{k=1}^K \) where $\boldsymbol{\tau}_{T}^k$ represents the text embeddings of the context prompt \( [\boldsymbol{P}_{sh}^k][\boldsymbol{P}_{T}][\text{CLASS}_k] \) for class \( k \). Let  
$\mathbb{P}_{\tau,\pi} = \sum_{k=1}^{K} \pi_{k} \delta_{\boldsymbol{\tau}_{T}^k}$
be the discrete distribution over the set of text embeddings \( \mathcal{T} \) for the target domain, where the category probabilities \( \pi \in \Delta_{K} = \{\alpha \ge 0 : \Vert \alpha \Vert_1 = 1 \} \) lie in the \( K \)-simplex. Additionally, let $\mathbb{P}^T = \frac{1}{N_T}\sum_{j=1}^{N_T}  \delta_{\boldsymbol{z}_j}$

represent the visual embedding distribution of the target domain, where $\delta$ is the Dirac delta  function and $\boldsymbol{z} = f_v(\boldsymbol{x})$ for the target image $\boldsymbol{x}$. The clustering assumption is then enforced by the following objective:
\begin{equation}
\mathcal{L}_{\mathcal{W}}=\mathcal{W}_{d_z}\left ( \mathbb{P}_{\tau,\pi}, \mathbb{P}^T\right) \label{eq:clustering}
\end{equation}
where $d_z$ represents a metric function. We use the cosine distance $d_z(a,b)=1-\langle  a,b \rangle$ since the visual embeddings and text embeddings already lie in the unit hypersphere. The following lemma demonstrates the behavior of $\pi$ and explains how the Wasserstein term helps target prompts enforce clustering properties.
\begin{lemma}
\label{cor:distortion_data} 
Let $\mathcal{T}^{*}=\left\{ \boldsymbol{\tau}_{T}^{k,*}\right\} _{k=1}^K$ be the optimal solution of the OP in Eq. (\ref{eq:clustering}), then $\mathcal{T}^{*}$ is also the optimal solution of the following OP:
\begin{equation}
\label{eq:distortion}
\min _{\mathcal{T},\pi}\min_{\sigma\in\Sigma_{\pi}}\mathbb{E}_{z\sim \mathbb{P}^T}\left [d_z\left(\boldsymbol{z},\tau^{\sigma(z)}_T\right) \right ],
\end{equation}
where $\Sigma_{\pi}$ is the set of assignment functions $\sigma:\mathcal{Z} \rightarrow\left\{ 1,...,K\right\} $
such that the cardinalities $ |\sigma^{-1}\left(k\right)|,k=1,...,K$
are proportional to $\pi_{k},k=1,...,K$. Moreover, given the set of text embeddings $\mathcal{T}$, the optimal $\sigma$ of the inner minimization is the nearest assignment: $\sigma^{-1}\left(k\right) = \{\boldsymbol{z}\mid k=\text{argmin} _{m}d_{z}\left(\boldsymbol{z},\boldsymbol{\tau}_{T}^{m}\right)\}$ is set of visual embeddings which are quantized to $k^{th}$ text embedding $\boldsymbol{\tau}^k_T$.
\end{lemma}

Lemma \ref{cor:distortion_data}~indicates the aim of minimizing the Wasserstein term $\mathcal{W}_{d_z}\left ( \mathbb{P}_{\tau,\pi}, \mathbb{P}^T\right)$, by adjusting the text embeddings in $\mathcal{T}$ to become the clustering centroids of the visual embeddings in the target domain, thereby minimizing the \textit{distortion between the text embeddings and target visual embeddings}. Specifically, we can interpret the optimization problem in (\ref{eq:distortion}) as finding the optimal clustering assignment $\sigma$, which maps each target example $\boldsymbol{z}$ to the text embedding $\boldsymbol{\tau}^{\sigma(z)}_T$ to minimize distortion. Moreover, given the text embeddings in $\mathcal{T}$, the inner minimization in (\ref{eq:distortion}) specifies the nearest assignment by assigning each visual embedding $\boldsymbol{z}$ to the closest text embedding. Furthermore, in the optimal solution, the optimal assignment $\sigma^*$, mapping target visual embeddings to clustering centroids (i.e., text embeddings), has a valuable property: \textit{the cardinalities $\left|(\sigma^*)^{-1}\left(k\right)\right|, k=1, \ldots, K$ are proportional to $\pi_{k}, k=1, \ldots, K$}. We set $\pi_k = \frac{1}{K}$ for $k = 1, \ldots, K$, aiming to learn balanced clusters.

It is important to note that the Wasserstein distance alone can shift the text embeddings toward the cluster centroids of the visual embedding distribution in the target domain. However, this may inadvertently cause the text embedding for one class (e.g., class 1) to shift to the centroid of a different class's cluster (e.g., class 2). To prevent this, pseudo-label guidance is necessary to ensure that each text embedding moves toward its correct cluster.

\paragraph{Final Objective.} We combine the objective functions for source prompts, target prompts, and cluster enforcement, and perform joint training as follows:
 \begin{align}
    \mathcal{L}_{total}(\boldsymbol{P}) := \mathcal{L}_{S}+ \lambda_T\mathcal{L}_{T}+ \lambda_\mathcal{W}\mathcal{L}_{\mathcal{W}}
\end{align}

where $\lambda_T$ and $\lambda_{\mathcal{W}}$ are hyper-parameters.
Here we note that in our experiments, we simply set $\lambda_T = \lambda_\mathcal{W} = 0.5$.

\section{Experiments}

In this section, we evaluate the efficacy of our proposed method on different UDA benchmarks, following the same protocol of recent prompt-based UDA studies \citep{ge2023domain, chen2022multiprompt, phan2024enhancing}. Their detailed descriptions are available in Supplementary.

\subsection{Experimental setup}


\noindent\textbf{Metrics.} Following \cite{phan2024enhancing}, we conduct experiments in two standard settings: a \textit{source-combined} setting, where data from all source domains are merged, and a \textit{multi-source} setting, which utilizes individual domain identifications. 

\noindent\textbf{Inference.}
During inference, unlike previous baselines (MPA \citep{chen2022multiprompt}, DAPL \citep{ge2023domain}, PGA \cite{phan2024enhancing}) that utilize both source and target domain prompts for predictions, we rely solely on the target prompt to evaluate whether we have effectively learned a meaningful textual description for the target domain. Further results, discussed in Section~\ref{sec:ablation_study}, reveal that previous methods fail to learn high-quality target prompts, as their performance heavily depends on the quality of source prompts. This dependency is obscured by the combined use of source and target prompts during prediction.

\subsection{Experimental results}

In the following results, our method is referred to as "Clustering Reinforcement Prompt Learning" (CRPL).

\begin{table}[h!]
\centering
\resizebox{1.\columnwidth}{!} {
\begin{tabular}{@{}lcccccccccc@{}}
\toprule
& \multicolumn{4}{c}{ImageCLEF} & \phantom{abc} & \multicolumn{4}{c}{Office-Home} \\
\cmidrule{2-5} \cmidrule{7-10}
& \textbf{$\rightarrow$ C} & \textbf{$\rightarrow$ I} & \textbf{$\rightarrow$ P} & \textbf{Avg} && \textbf{$\rightarrow$ Ar} & \textbf{$\rightarrow$ Cl} & \textbf{$\rightarrow$ Pr} & \textbf{$\rightarrow$ Rw} & \textbf{Avg} \\
\midrule
\textbf{Zero-Shot} &&&&&&&&& \\
CLIP \citep{radford2021learning} & 87.9 & 88.2 & 78.7 & 88.1 && 71.2 & 50.4 & 81.4 & 82.6 & 71.4 \\
\midrule
\textbf{Source Combined} &&&&&&&&& \\
DAN \citep{ganin2015unsupervised} & 93.3 & 92.2 & 77.6 & 87.7 && 68.5 & {59.4} & 79.0 & 82.5 & 72.4 \\
DANN \citep{ganin2016domain} & 93.7 & 91.8 & 77.9 & 87.8 && 68.4 & 59.1 & 79.5 & 82.7 & 72.4 \\
D-CORAL \citep{sun2016deep} & 93.6 & 91.7 & 77.1 & 87.5 && 68.1 & 58.6 & 79.5 & 82.7 & 72.2 \\
DAPL \citep{ge2023domain} & 96.0 & 89.2 & 76.0 & 87.1 && 72.8 & 51.9 & 82.6 & 83.7 & 72.8 \\
Simple Prompt \citep{chen2022multiprompt} & 93.6 & 90.6 & 80.9 & 88.4 && 70.7 & 52.9 & 82.9 & 83.9 & 72.4 \\
PGA \cite{phan2024enhancing} & 94.2 & 92.1 & 78.5 & 88.2 &&  74.1 & 53.9 & 84.4 & 85.6 & 74.5\\
\rowcolor[gray]{0.9}CRPL (Ours) & \textbf{94.8} & \textbf{94.5} & \textbf{81.7} & \textbf{90.3} &&  {\textbf{76.6}} & \textbf{60.4} & {\textbf{86.5}} & {\textbf{86.8}} & {\textbf{77.6}}\\
\midrule
\textbf{Multi-Source} &&&&&&&&& \\
DCTN \citep{xu2018deep} & 95.7 & 90.3 & 75.0 & 87.0 && N.A. & N.A. & N.A. & N.A. & N.A. \\
MDDA \citep{DBLP:conf/aaai/ZhaoWZGLS0HCK20} & N.A. & N.A. & N.A. & N.A. && 66.7 & {62.3} & 79.5 & 79.6 & 71.0 \\
SIMplDA \citep{venkat2021classifier} & 93.3 & 91.0 & 77.5 & 87.3 && 70.8 & 56.3 & 80.2 & 81.5 & 72.2 \\
MFSAN \citep{DBLP:conf/aaai/ZhuZW19} & 95.4 & 93.6 & 79.1 & 89.4 && 72.1 & 62.0 & 80.3 & 81.8 & 74.1 \\
MPA \citep{chen2022multiprompt}   & \textbf{97.2} & \textbf{96.2} & 80.4 & 91.3 && 74.8 & 54.9 & {86.2} & 85.7 & 75.4 \\
MPGA \cite{phan2024enhancing}  & 93.8  & 95.7 & \textbf{82.8} & 90.8 & & 74.8 & 56.0 & 85.2 & 86.0 & 75.5 \\
\rowcolor[gray]{0.9}M-CRPL (Ours) & 96.2 & 96.0 & {82.3} & \textbf{91.5} & & \textbf{76.8} & \textbf{63.5} & \textbf{87.6} & \textbf{87.5} & \textbf{78.9} \\
\bottomrule
\end{tabular}}
\vspace{-2mm}
\caption{Accuracy (\%) on ImageCLEF and Office-Home. We use \textbf{bold} to denote the best method overall.\label{tab:imageclef}}
\vspace{-3mm}
\end{table}
Table \ref{tab:imageclef} presents the results for the ImageCLEF and Office-Home datasets. Under the source-combined scenario, 
our method significantly outperforms all other baseline methods on both datasets and target domain settings. For instance, CRPL surpasses the second-best source combined method by $1.9\%$ in average accuracy for ImageCLEF and $3.1\%$ in average accuracy for Office-Home. In the multi-source setting, our method delivers the highest average performance across dataset, notably outperforming MPGA, the state-of-the-art (SOTA) prompt-based method for multi-source UDA, by a substantial margin of $3.4\%$ on Office-Home.
\begin{table}[!ht]
\centering
\resizebox{1.\columnwidth}{!} {
\begin{tabular}{@{}lcccccccc@{}}
\toprule
& \multicolumn{7}{c}{DomainNet} & \\
\cmidrule{2-8}
& \textbf{$\rightarrow$ Clp} & \textbf{$\rightarrow$ Inf} & \textbf{$\rightarrow$ Pnt} & \textbf{$\rightarrow$ Qdr} & \textbf{$\rightarrow$ Rel} & \textbf{$\rightarrow$ Skt} & \textbf{Avg} \\
\midrule
\textbf{Zero-Shot} &&&&&&& \\
CLIP \citep{radford2021learning} & 61.3 & 42.0 & 56.1 & 10.3 & 79.3 & 54.1 & 50.5 \\
\midrule
\textbf{Source Combined} &&&&&&& \\
DANN \citep{ganin2016domain} & 45.5 & {13.1} & 37.0 & \textbf{13.2} & 48.9 & 31.8 & 32.6 \\
MCD \citep{saito2017maximum} & 54.3 & 22.1 & 45.7 & 7.6 & 58.4 & 43.5 & 38.5 \\
DAPL \citep{ge2023domain} & 62.4 & 43.8 & 59.3 & 10.6 & {81.5} & 54.6 & 52.0 \\
Simple Prompt \citep{chen2022multiprompt}  & 63.1 & 41.2 & 57.7 & 10.0 & 75.8 & 55.8 & 50.6 \\
PGA \cite{phan2024enhancing} & {65.4} & 49.0 & 60.4 & {11.1} & \textbf{81.8} & {60.6} & {55.4} \\
\rowcolor[gray]{0.9}CRPL (Ours) & \textbf{65.6} & \textbf{50.8} & \textbf{66.7}  & 10.6 & 80.0 & \textbf{61.1} & \textbf{55.8} \\
\midrule
\textbf{Multi-Source} &&&&&&& \\
M³SDA-$\beta$ \citep{DBLP:conf/iccv/PengBXHSW19} & 58.6 & 26.0 & 52.3 & 6.3 & 62.7 & 49.5 & 42.6 \\
SImpAl101 \citep{venkat2021classifier} & 66.4 & 26.5 & 56.6 & \textbf{18.9} & 68.0 & 55.5 & 48.6 \\
LtC-MSDA \citep{DBLP:conf/eccv/WangXN020} & 63.1 & 28.7 & 56.1 & 16.3 & 66.1 & 53.8 & 47.4 \\
T-SVDNet \citep{li2021tsvdnet} & 66.1 & 25.0 & 54.3 & 16.5 & 65.4 & 54.6 & 47.0 \\
PFSA \citep{Fu_2021_CVPR} & 64.5 & 29.2 & 57.6 & {17.2} & 67.2 & 55.1 & 48.5 \\
PTMDA \citep{ren2022multisource}  & 66.0 & 28.5 & 58.4 & 13.0 & 63.0 & 54.1 & 47.2 \\
MPA \citep{chen2022multiprompt}  & 65.2 & 47.3 & 62.0 & 10.2 & \textbf{82.0} & 57.9 & 54.1 \\
MPGA \cite{phan2024enhancing} & {67.2} & {47.8} & {63.1} & 11.6 & {81.7} & \textbf{61.0 }& {55.4} \\
\rowcolor[gray]{0.9}M-CRPL (Ours) & \textbf{67.6} & \textbf{51.4} & \textbf{67.0} & 11.1 & 79.3 & {60.6} &  \textbf{56.2}\\
\bottomrule
\end{tabular}}
\vspace{-2mm}
\caption{Accuracy (\%) on DomainNet. We use \textbf{bold} to denote the best method overall.\label{tab:domainnet_tab}}
\end{table}

\begin{table*}[!h]
\centering
\resizebox{1\textwidth}{!}{
    \begin{tabular}{lccccccccccccc}
    \toprule
        Method & Ar→Cl & Ar→Pr & Ar→Rw & Cl→Ar & Cl→Pr & Cl→Rw & Pr→Ar & Pr→Cl & Pr→Rw & Rw→Ar & Rw→Cl & Rw→Pr & Avg  \\
        \midrule
        ResNet-50\cite{he2016deep} & 34.9 & 50.0 & 58.0 & 37.4 & 41.9 & 46.2 & 38.5 & 31.2 & 60.4 & 53.9 & 41.2 & 59.9 & 46.1  \\ 
        DANN \cite{ganin2015unsupervised} & 45.6 & 59.3 & 70.1 & 47.0 & 58.5 & 60.9 & 46.1 & 43.7 & 68.5 & 63.2 & 51.8 & 76.8 & 57.6  \\ 
        JAN \cite{long2017deep} & 45.9 & 61.2 & 68.9 & 50.4 & 59.7 & 61.0 & 45.8 & 43.4 & 70.3 & 63.9 & 52.4 & 76.8 & 58.3 \\
        
        CDAN+E \cite{long2017conditional} & 50.7 & 70.6 & 76.0 & 57.6 & 70.0 & 70.0 & 57.4 & 50.9 & 77.3 & 70.9 & 56.7 & 81.6 & 65.8 \\ 
        BSP+CDAN \cite{BSP_ICML2019} & 52.0 & 68.6 & 76.1 & 58.0 & 70.3 & 70.2 & 58.6 & 50.2 & 77.6 & 72.2 & 59.3 & 81.9 & 66.3 \\ 
        SymNets \cite{zhang2019domain} & 47.7 & 72.9 & 78.5 & 64.2 & 71.3 & 74.2 & 63.6 & 47.6 & 79.4 & 73.8 & 50.8 & 82.6 & 67.2  \\
        
        ETD \cite{ETD_CVPR20} & 51.3 & 71.9 & \textbf{85.7} & 57.6 & 69.2 & 73.7 & 57.8 & 51.2 & 79.3 & 70.2 & 57.5 & 82.1 & 67.3 \\
       
        BNM \cite{BNM_CVPR2020} & 52.3 & 73.9 & 80.0 & 63.3 & 72.9 & 74.9 & 61.7 & 49.5 & 79.7 & 70.5 & 53.6 & 82.2 & 67.9 \\ 
        
        GSDA \cite{hu2020unsupervised} & {\textbf{61.3}} & 76.1 & 79.4 & 65.4 & 73.3 & 74.3 & 65.0 & 53.2 & 80.0 & 72.2 & {\textbf{60.6}} & 83.1 & 70.3  \\ 
        GVB-GD \cite{cui2020gradually} & 57.0 & 74.7 & 79.8 & 64.6 & 74.1 & 74.6 & 65.2 & {55.1} & 81.0 & 74.6 & 59.7 & 84.3 & 70.4  \\ 
        RSDA-MSTN \cite{gu2020spherical} & 53.2 & 77.7 & 81.3 & 66.4 & 74.0 & 76.5 & 67.9 & 53.0 & 82.0 & 75.8 & 57.8 & 85.4 & 70.9  \\ 
        SPL \cite{SPL_AAAI20} & 54.5 & 77.8 & 81.9 & 65.1 & 78.0 & 81.1 & 66.0 & 53.1 & 82.8 & 69.9 & 55.3 & {\textbf{86.0}} & 71.0 \\
       
        SRDC \cite{tang2020unsupervised} & 52.3 & 76.3 & 81.0 & 69.5 & 76.2 & 78.0 & 68.7 & 53.8 & 81.7 & \textbf{{76.3}} & 57.1 & 85.0 & 71.3  \\ 
        DisClusterDA \citep{tang2022unsupervised}& 58.8 &77.0& 80.8& 67.0& 74.6& 77.1& 65.9& \textbf{56.3}& 81.4& 74.2& 60.5& 83.6& 71.4 \\
        \bottomrule
        CLIP \cite{radford2021learning} & 51.6 & 81.9 & 82.6 & 71.9 & 81.9 & 82.6 & 71.9 & 51.6 & 82.6 & 71.9 & 51.6 & 81.9 & 72.0  \\ 
DAPL \citep{ge2023domain} & 52.7 & 82.2 & 84.1 & 73.9 & 82.0 & \textbf{83.8} & 73.6 & 54.6 & 84.0 & 73.3 & 53.4 & 82.5  & 73.3\\ 
PGA \cite{phan2024enhancing}  &53.7 & 83.9 & 85.0 & 73.2 & 83.9  & 84.6  & 73.2  &  53.8& 84.1 & 73.5  &  53.1& 85.3 & 73.9\\
\rowcolor[gray]{0.9} CRPL (Ours)  & 54.7 & \textbf{84.1} & 84.6 & \textbf{74.3}  &\textbf{83.2} & {83.7} & \textbf{73.7} & 53.4 & \textbf{84.6} & {74.5}  & 55.5 & 85.5 & \textbf{74.4}
        \\
        \bottomrule
    \end{tabular}}
    \vspace{-2mm}
    \caption{Accuracy (\%) on Office-Home\cite{Office-Home} for unsupervised domain adaptation (ResNet-50\cite{he2016deep}). The best accuracy is indicated in \textbf{bold}.}
    \label{tab:officehome}
    \label{tab:officehome_pair}
\end{table*}
On DomainNet, as shown in Table~\ref{tab:domainnet_tab}, we achieve superior average accuracy under both source-combined and multi-source settings, boosting the performance of pretrained CLIP by $\approx6\%$. We also consistently outperform other baselines on the (`Clp', `Inf', and `Pnt') domains across all settings, while showing competitive performance in the other three domains. This is because domains such as (`Clp', `Inf', and `Pnt') exhibit highly similar characteristics, whereas (`Qdr', `Rel', and `Skt') are significantly different, which impacts transferability from other domains. For more distinct domains, reference labels from the base prompts may be more appropriate. However, we do not have access to the data used to train the base prompts, which limits our ability to establish the relationship between the target domain and the base prompts. This makes it difficult to decide when to use the base prompts or the source-enhanced prompts. As a result, we apply an average combination of the base prompt and the source-enhanced prompt, as shown in Eq.~(\ref{eq:enhance_prompt}), to balance their contributions.

The effectiveness of our approach is further demonstrated through 12 pairwise source-target settings on Office-Home, as shown in Table \ref{tab:officehome_pair}. While prompt-based methods using pre-trained CLIP outperform traditional UDA baselines, CRPL achieves the highest average score and consistently outperforms both DAPL and PGA across all 12 settings.

\subsection{Ablation study}
\label{sec:ablation_study}

In this section, we conduct several ablation studies on the Office-Home dataset to evaluate the effectiveness of the two key components of our proposed approach: \textit{source-enhanced pseudo-labels} and \textit{clustering reinforcement}, denoted as \( \mathcal{L}_\mathcal{W} \). We also analyze the impact of the \textit{source-target prompt alignment} strategy in previous approaches. In the following, \textbf{CPL} refers to using hard labels obtained from CLIP’s zero-shot predictions to train the target prompt; \textbf{SPL} refers to using enhanced
pseudo-labels derived from the source domains to train the target prompt; and \textbf{with $\mathcal{L}_{\mathcal{W}}$} refers to adding the optimal transport (OT) clustering to enhance performance. Additionally, we consider different scenarios, including using only the target prompt ($\tau_{T}$), averaging the source prompts ($\tau_{S}$), and averaging all prompts ($\tau_{avg}$) for inference.

\begin{table}[h!]
\centering
\resizebox{0.7\columnwidth}{!}{\begin{tabular}{@{}llcccc@{}}
\toprule
& Inference Prompt & Ar & Cl & Pr & Rw \\
\midrule
& $\tau_{T}$ &47.6 & 29.0 & 53.5  & 63.5\\
CPL & $\tau_{S}$ & 74.3 & 57.2 & 84.3 &84.8 \\
& $\tau_{avg}$ & 61.2 & 40.3 &73.3 & 77.8  \\
\midrule
& $\tau_{T}$ & 75.8 & 62.9 & 86.6 & 87.2\\
SPL& $\tau_{S}$ &73.3 & 57.4 & 83.4 & 84.7 \\
& $\tau_{avg}$ &75.8 & 58.1 & 83.6 & 85.3 \\
\midrule
\midrule
CPL only & $\tau_{T}$ &47.6 & 29.0 & 53.5  & 63.5\\
CPL with $\mathcal{L}_\mathcal{W}$& $\tau_{T}$ &7.9 & 4.7 & 80.4  & 82.3\\
\midrule
SPL only & $\tau_{T}$ & 75.8 & 62.9 & 86.6 & 87.2 \\
SPL with $\mathcal{L}_\mathcal{W}$  & $\tau_{T}$ & 76.8 & 63.5 & 87.5 & 87.6 \\
\bottomrule
\end{tabular}
}
\vspace{-2mm}
\caption{Effect of source-enhanced
pseudo-labels and clustering reinforcement on the Office-Home dataset.\label{tab:ablation_pseudo_label}}
\vspace{-3mm}
\end{table}

\noindent\textbf{Effect of SPL.} Table~\ref{tab:ablation_pseudo_label} demonstrates that SPL significantly improves the performance of CPL. Moreover, when comparing performance across different inference scenarios ($\tau_{T}$, $\tau_{S}$, $\tau_{avg}$), it becomes evident that source-target prompt alignment in the simple baseline has a limited impact on the target prompt, as evidenced by the significantly lower accuracy in $\tau_{T}$ compared to $\tau_{S}$. In contrast, SPL enables the target domain to learn text embeddings that effectively capture and represent the corresponding visual embeddings.


\noindent\textbf{Effect of clustering reinforcement \( \mathcal{L}_\mathcal{W} \).} Table~\ref{tab:ablation_pseudo_label} (last four rows) shows that clustering reinforcement \( \mathcal{L}_\mathcal{W} \) further improves the performance of SPL. However, applying \( \mathcal{L}_\mathcal{W} \) to the simple baseline reveals noticeable differences in CPL performance over target domains. 
Specifically, it seems that the effectiveness of $\mathcal{L}_\mathcal{W}$ strongly depends on the initial predictions, where weak initial predictions as observed in the CPL method on `Ar' and `Cl' domains with just 47.6\% and 29.0\%, leading to a collapse performance of accuracy less than 8\% when combing with $\mathcal{L}_\mathcal{W}$. However, with better initial predictions as observed in the CPL method on `Pr' and `Rw' domains, substantial improvement of 27\% (53.5\% to 80.4\%) and 19\% (63.5\% to 82.3\%) when combined with $\mathcal{L}_\mathcal{W}$ are observed. This indicates a mutual benefit between source-enhanced
pseudo-labels and clustering reinforcement.

\begin{table}[ht]
\centering
\resizebox{0.7\columnwidth}{!}{\begin{tabular}{@{}llcccc@{}}
\toprule
Setting & Inference Prompt & Ar & Cl & Pr & Rw \\
\midrule
PGA & $\tau_{T}$ &72.3 & 52.8 & 83.7  & 82.4 \\
  & $\tau_{S}$  &73.8 & 56.2 & 84.6 &84.9 \\
  & $\tau_{avg}$  &75.6 & 57.0 & 86.0 &86.6 \\
  \midrule
M-CRPL (ours)  & $\tau_{T}$ & \textbf{76.8} & \textbf{63.5} & \textbf{87.5} & \textbf{87.6} \\
  & $\tau_{S}$ & 75.3 & 58.0 & 83.9& 84.7 \\
 & $\tau_{avg}$ & 75.8 &58.6 &85.2 & 85.3\\
\bottomrule
\end{tabular}
}
\vspace{-2mm}
\caption{Analyses of prompts performance on OfficeHome dataset.}
\label{tab:aligment_effect}
\vspace{-3mm}
\end{table}
\noindent\textbf{Effect of Source-Target Prompts Alignment.} We have shown that source-target prompt alignment has a limited impact on the target prompt. To further validate this, we analyze the performance of prompts from the most recent SOTA method, PGA \cite{phan2024enhancing}, which focuses on the source-target prompt strategy. As shown in Table~\ref{tab:aligment_effect}, while PGA improves the performance of \( \tau_T \), bringing it closer to the performance of the source prompts \( \tau_S \), it does not surpass them. This result confirms the limitations of the alignment strategy. In contrast, for M-CRPL, \( \tau_T \) achieves the best performance for target prompts across all three scenarios, demonstrating the effectiveness of using enhanced pseudo-labels from source domains to refine target prompts in terms of representation.



\section{Conclusion}
We propose a framework for prompt learning UDA that begins by leveraging information from the source domains to enhance pseudo-labels for training target prompts. We demonstrate that while source-target prompt alignment has a limited effect on learning the target prompt, enhanced pseudo-labels result in significant improvements in both performance and the quality of target prompts in terms of representation. Additionally, inspired by the strong clustering behavior observed between visual and text embeddings in pre-trained models, we further refine target prompts by enforcing clustering properties. This is achieved by minimizing the Wasserstein distance between the distribution of text prompts and the distribution of visual embeddings in the target domains. Finally, our experiments and ablation studies validate the superior performance of our approach.

\section*{Acknowledgements}
This work was supported by ARC DP23 grant DP230101176 and by the Air Force Office of Scientific Research under award number FA2386-23-1-4044.


{
    \small
    \bibliographystyle{ieeenat_fullname}
    \bibliography{main}

}

\clearpage
\newpage
\mbox{}

\section*{Appendix}


In this supplementary material, we present the experimental settings and implementation details in Section~\ref{sec:setting}, followed by additional discussions and experiments in Section~\ref{sec:observation}. The limitations of our current work are outlined in Section~\ref{sec:limitation}, and the proof of Lemma 1 is provided in Section~\ref{sec:proof}.

\section{Experimental Settings}
\label{sec:setting}

\paragraph{Datasets}
\label{sec:supp_datasets}
ImageCLEF is a small-scaled dataset with 1,800 images across 12 object categories from three domains: ImageNet ILSVRC 2012 (I), Pascal VOC 2012 (P), and Caltech-256 (C). Office-Home is a medium-scaled dataset containing approximately 15,500 images from 65 categories in four domains: Art, Clipart, Product, and Real World. DomainNet is the largest dataset, comprising around 600,000 images from 345 categories across six domains: Clipart, Infograph, Painting, Quickdraw, Real, and Sketch.

\paragraph{Baselines.} Regarding prompt-based baselines, we compare our method with MPA \citep{chen2022multiprompt}, DAPL \citep{ge2023domain}, Simple Prompt \citep{chen2022multiprompt}, PGA \cite{phan2024enhancing}, and Zero-shot CLIP \citep{radford2021learning}. To ensure a comprehensive evaluation, we also include comparisons with various non-prompt methods such as DCTN \citep{xu2018deep}, MDDA \citep{DBLP:conf/aaai/ZhaoWZGLS0HCK20}, MFSAN \citep{DBLP:conf/aaai/ZhuZW19}, T-SVDNet \citep{li2021tsvdnet}, and PFSA \citep{Fu_2021_CVPR}. As we follow the same settings as in \citep{chen2022multiprompt, ge2023domain, phan2024enhancing}, the results for these baselines are reproduced based on their public implementations and hyper-parameters to ensure consistency. Note that while DAPL, MPA, PGA, and our method employ CoOp \citep{zhou2022learning} with text-end soft prompts, other methods fine-tune the transformer block \citep{du2024domain}, both image and text-end soft prompts \citep{lai2024empowering}, or the entire encoders \citep{lai2023padclip, zhou2024unsupervised}. Since these alternative methods typically fine-tune many more parameters, we exclude them from the experiments to ensure a fair comparison.

\paragraph{Metrics.} We use the top-1 accuracy for each target domain and the average accuracy across all domains as the evaluation metric. Following \cite{phan2024enhancing}, we conduct experiments in two standard settings: a \textit{source-combined} setting, where data from all source domains are merged, and a \textit{multi-source} setting, which utilizes individual domain identifications. We also provide pair-wise single-source domain adaptation results for the Office-Home dataset. 


\paragraph{Implementation details} 
For fair comparisons, we use ResNet50 as our backbone on the Image-CLEF and Office-Home datasets and ResNet101 on DomainNet. The weights are initialized from a pre-trained CLIP model and kept frozen during training. Following previous baselines \cite{ge2023domain, chen2022multiprompt, phan2024enhancing}, prompts are trained using the mini-batch SGD optimizer with a learning rate of 0.005. We use a batch size of 32 and apply a cosine learning rate scheduler. For hyperparameters, token lengths \( M_1 \) and \( M_2 \) are both set to 16. Additionally, we do not require a pseudo-label threshold \( \tau \) for label generation as we combine base-prompt and source-prompt to generate enhanced soft pseudo-labels.
For our specific parameter $\lambda_T$ and $\lambda_\mathcal{W}$, we simply set $\lambda_T=\lambda_\mathcal{W}=0.5$ for all experiments. \textbf{Code available at}: \url{https://github.com/VuongLong/Clustering-Reinforcement-Prompt-Learning/}






\section{Additional Experiments}
\label{sec:observation}

\subsection{Distance-Aware Pseudo-Label}
As discuss in previous section, different transferability between domains motivate us a distance aware pseudo-labels scheme. Specifically, we calculate the weighted average cosine distance from the visual embedding $\boldsymbol{z}=f_v(\boldsymbol{x})$ to the text embeddings of each class across all source domains. We recap Eq.~(\ref{eq:enhance_prompt}) in the main for easier readability:

\begin{equation}
   \boldsymbol{\tau}_{ave}^k(x) = \frac{1}{2}\boldsymbol{\tau}_{base}^k  + \frac{w_{k,i}(x)}{2}\sum_{i=1}^N \boldsymbol{\tau}_{S_i}^k
   \label{eq:enhance_prompt_apd}
\end{equation}

and $w_{i,k}$ is the important weight for each domain-class. 

Here we use the \( l_2 \) distance between the visual embedding \( z^{pre} \) (the unnormalized visual embedding of \( x \), where \( z = \frac{z^{pre}}{\| z^{pre} \|_2} \)) and the centroid of class \( k \) in domain \( i \). Let \( C^i = \{c^i_k\}_{k=1}^K \) represent the set of class centroids for the \( i \)-th domain i.e., $c^i_k = \frac{1}{\sum_{j=1}^{N_{S_i}}\mathbb{I}_{(y^i_j=k)}}\sum_{j=1}^{N_{S_i}}\mathbb{I}_{(y^i_j=k)}  \boldsymbol{z}^{pre,i}_j \nonumber$. The weight for class \( k \) is then calculated by applying the softmax function:

\begin{equation}
    w_{i,k}(z) = \frac{\exp\left ( \left \| z_{pre}-c^i_k \right \|_2\right )}{\sum_{i'=1}^N\exp\left(\left \| z-c^{i'}_k \right \|_2\right )}
\end{equation}

It is important to highlight that the weights \(w_{i,k}(z)\) are applied to compute the combined prompts \(\boldsymbol{\tau}_{\text{ave}}^k\), rather than being applied directly to the outputs of the softmax (i.e., the predictions from individual prompts). Specifically, the enhanced pseudo-labels are computed as described in Eq.~(\ref{eq:enhanced_label}) in the main paper:
\begin{equation}
    \hat{y}[k] = \frac{\exp\left(\langle \boldsymbol{z}, \boldsymbol{\tau}_{ave}^k((x)) \rangle/\gamma\right )}{\sum_{k'=1}^K\exp\left(\langle \boldsymbol{z}, \boldsymbol{\tau}_{ave}^{k'}(x) \rangle/\gamma\right )}
\end{equation}
where 
\begin{equation}
    \langle \boldsymbol{z}, \boldsymbol{\tau}_{ave}^k((x)) \rangle=\frac{1}{2} \langle \boldsymbol{z}, \boldsymbol{\tau}_{base}^k\rangle  + \frac{w_{k,i}(x)}{2}\sum_{i=1}^N  \langle \boldsymbol{z}, \boldsymbol{\tau}_{S_i}^k\rangle
\end{equation}
It can be observed that this design preserves semantic similarity, as the magnitude of cosine similarity between the visual embedding and the text embedding of each individual prompt is taken into account in the final pseudo-label. Meanwhile, the weights \(w_{k,i}(x)\) represent the spatial relationship between the visual embedding and the text embedding.

\begin{table}[ht]
\centering
\resizebox{0.8\columnwidth}{!}{\begin{tabular}{@{}llcccc@{}}
\toprule
Distance  & Ar & Cl & Pr & Rw & Average \\
\midrule
Average & 76.0 & 62.6 & 87.0 & 87.5 & 78.3\\
cosine & 76.6 & 56.2 & 88.2 & 86.7 & 76.9\\
L2 & \textbf{76.8} & \textbf{63.5} & \textbf{87.5} & \textbf{87.6} & \textbf{78.9}\\
\bottomrule
\end{tabular}
}
\caption{Different metric for Distance-aware Pseudo-Labels on Office-Home dataset.}
\label{tab:distance}
\end{table}

To evaluate the effectiveness of this design, we compare the performance of the proposed distance-aware pseudo-labeling scheme with two alternative strategies: a simple averaging of source prompts and a distance-aware approach using cosine distance instead of \(L_2\). For the cosine distance, we use \(z\) rather than \(z_{\text{pre}}\) to compute weights and class centroids. As shown in Table~\ref{tab:distance}, the \(L_2\) distance achieves the best performance, while the cosine distance performs the worst. This discrepancy may be due to overlapping information between the cosine similarity of visual embeddings with text embeddings of prompts and the cosine similarity of visual embeddings with class centroids.
\subsection{Performance with ViT models} 

\begin{table}[h!]
\centering
\resizebox{0.7\columnwidth}{!}{\begin{tabular}{@{}llcccc@{}}
\toprule
Setting  & Ar & Cl & Pr & Rw & Ave \\
\midrule
Zero-Shot &86.6 & 72.4 & 92.8  & 92.7 &86.1\\
\midrule
PGA &87.5 & 77.2 & 94.5  & 94.5 &88.4\\
  \midrule
ours   & \textbf{89.7} & \textbf{83.2} & \textbf{95.9} & \textbf{95.5} & \textbf{91.1} \\
\bottomrule
\end{tabular}
}
\caption{Performance of ViT-14L on OfficeHome dataset.}
\label{tab:VIT}
\end{table}

In the main paper, we conduct experiments using the ResNet backbone. Additionally, we perform experiments with the ViT-14L model. The results presented in Table~\ref{tab:VIT}, comparing our approach with the most recent SOTA method, PGA, demonstrate that our approach outperforms previous methods.

\subsection{Performance of corrupted dataset}

To further assess the robustness of our method, we performed experiments using corrupted data generated from the Office-Home dataset (for details on the corruptions, refer to \url{https://github.com/hendrycks/robustness}).

\begin{table}[h!]
\centering
\resizebox{1.0\columnwidth}{!}{\begin{tabular}{@{}llccccc@{}}
\toprule
Corruption &Method & Ar & Cl & Pr & Rw & Ave \\
\midrule
No corruption &Zero-Shot &71.2 &50.4 &81.4& 82.6& 71.4\\
&PGA  &74.8 & 56.0 &85.2& 86.0& 75.5\\
&ours   & \textbf{76.8} & \textbf{63.5} & \textbf{87.5} & \textbf{87.6} & \textbf{78.9}\\
\midrule
Defocus\_blur &Zero-Shot &60.2 & 40.7 & 73.8  & 76.8 & 62.9\\
&PGA  &66.2 & 50.0 & 77.6 & 80.3 &68.5\\
&ours   & \textbf{69.1} & \textbf{61.3} & \textbf{78.1} & \textbf{81.1} & \textbf{72.4}\\
\midrule

elastic\_transform &Zero-Shot &635 & 411 & 74.2  & 77.3 & 64.0\\
&PGA  &69.4 & 48.1 & 80.7 & 82.3 &70.1\\
&ours   & \textbf{69.3} & \textbf{60.5} & \textbf{82.1} & \textbf{83.2} & \textbf{73.8}\\
\midrule

Gaussian\_noise &Zero-Shot &62.0 & 52.8 & 73.7  & 75.2 & 65.9\\
&PGA  &67.3 & 48.9 & \textbf{75.8}  & 79.2 &67.8\\
&ours   & \textbf{68.1} & \textbf{60.2} & {75.2} & \textbf{80.6} & \textbf{71.0}\\

\midrule
Speckle\_npose &Zero-Shot &62.1 & 38.5 & 65.3  & 71.2 & {59.2}\\
&PGA  &67.3 & 47.0 & \textbf{74.2}  & 78.3 & 66.7\\
&ours   & \textbf{67.7} & \textbf{58.0} & \textbf{73.7} & \textbf{78.8} & \textbf{69.5}\\
\bottomrule
\end{tabular}
}
\caption{Performance on Corrupted OfficeHome dataset.}
\label{tab:corrupted}
\end{table}

The results in Table~\ref{tab:corrupted} demonstrate that our method remains robust even as CLIP's zero-shot performance declines.

\section{Limitations}
\label{sec:limitation}
One of our main contributions is the enhancement of pseudo-labels for target domains by effectively leveraging information from source domains. Specifically, we analyze the relationship between the target and source domains, considering both semantic similarity (via cosine distance between visual embeddings and text embeddings of prompts) and spatial relationships (via \(L_2\) distance in the pre-normalized embedding space), to appropriately weight references from source domains. Additionally, we demonstrate the equivalence between the references provided by the base prompt and those from source prompts to the target prompt. However, due to the lack of access to the data used for training the base prompt, we are unable to directly compare the quality of references between the base prompt and source prompts. This limitation results in suboptimal utilization of these references, as we currently assign equal weight to the base prompt and the weighted source prompts to balance their contributions, as described in Eq.~(\ref{eq:enhance_prompt_apd}). Addressing this limitation is a key focus for future work.

\section{Proof of Lemma 1}
\label{sec:proof}
We propose minimizing the Wasserstein distance between the target prompts' text embeddings and the visual embeddings from the target domain. Specifically, denote \( \mathcal{T} = \{\boldsymbol{\tau}_{T}^k\}_{k=1}^K \) where $\boldsymbol{\tau}_{T}^k$ represents the text embeddings of the context prompt \( [\boldsymbol{P}_{sh}^k][\boldsymbol{P}_{T}][\text{CLASS}_k] \) for class \( k \). Let  
\begin{equation}
    \mathbb{P}_{\tau,\pi} = \sum_{k=1}^{K} \pi_{k} \delta_{\boldsymbol{\tau}_{T}^k}
\end{equation}
be the discrete distribution over the set of text embeddings \( \mathcal{T} \) for the target domain, where the category probabilities \( \pi \in \Delta_{K} = \{\alpha \ge 0 : \Vert \alpha \Vert_1 = 1 \} \) lie in the \( K \)-simplex. Additionally, let 
\begin{equation}
\mathbb{P}^T = \frac{1}{N_T}\sum_{j=1}^{N_T}  \delta_{\boldsymbol{z}_j} \nonumber
\end{equation}
represent the visual embedding distribution of the target domain, where $\delta$ is the Dirac delta  function and $\boldsymbol{z} = f_v(\boldsymbol{x})$ for the target image $\boldsymbol{x}$. The clustering assumption is then enforced by the following objective:
\begin{equation}
\mathcal{L}_{\mathcal{W}}=\mathcal{W}_{d_z}\left ( \mathbb{P}_{\tau,\pi}, \mathbb{P}^T\right) \label{eq:clustering_apd}
\end{equation}

where $d_z$ represents a metric function. We use the cosine distance $d_z(a,b)=1-\langle  a,b \rangle$ since the visual embeddings and text embeddings already lie in the unit hypersphere. The following lemma demonstrates the behavior of $\pi$ and explains how the Wasserstein term helps target prompts enforce clustering properties.

\begin{lemma} \textbf{(Lemma 1 in the main paper)}
\label{cor:distortion_data_apd} 
Let $\mathcal{T}^{*}=\left\{ \boldsymbol{\tau}_{T}^{k,*}\right\} _{k=1}^K$ be the optimal solution of the OP in Eq. (\ref{eq:clustering_apd}), then $\mathcal{T}^{*}$ is also the optimal solution of the following OP:
\begin{equation}
\label{eq:distortion}
\min _\mathcal{T,\pi}\min_{\sigma\in\Sigma_{\pi}}\mathbb{E}_{z\sim \mathbb{P}^T}\left [d_z\left(\boldsymbol{z},\tau^{\sigma(z)}_T\right) \right ],
\end{equation}
where $\Sigma_{\pi}$ is the set of assignment functions $\sigma:\mathcal{Z} \rightarrow\left\{ 1,...,K\right\} $
such that the cardinalities $ |\sigma^{-1}\left(k\right)|,k=1,...,K$
are proportional to $\pi_{k},k=1,...,K$. Moreover, given the set of text embeddings $\mathcal{T}$, the optimal $\sigma$ of the inner minimization is the nearest assignment: $\sigma^{-1}\left(k\right) = \{\boldsymbol{z}\mid k=\text{argmin} _{m}d_{z}\left(\boldsymbol{z},\boldsymbol{\tau}_{T}^{m}\right)\}$ is set of visual embeddings which are quantized to $k^{th}$ text embedding $\boldsymbol{\tau}^k_T$.
\end{lemma}

\noindent\textit{Proof:}

It is clear that
\[
\mathbb{P}_{\tau,\pi}=\sum_{k=1}^{K}\pi_{k}\delta_{\tau_T^{k}}.
\]

Therefore, we reach the following OP:
\begin{equation*}
\min_{\mathcal{T},\pi}\mathcal{W}_{d_{z}}\left(\frac{1}{N}\sum_{n=1}^{N_T}\delta_{z_{n}},\sum_{k=1}^{K}\pi_{k}\delta_{\tau_T^{k}}\right).\label{eq:appendix_push_forward-1}
\end{equation*}

By using the Monge definition, we have 
\begin{align*}
&\mathcal{W}_{d_{z}}\left(\frac{1}{N_T}\sum_{n=1}^{N_T}\delta_{z_{n}},\sum_{k=1}^{K}\pi_{k}\delta_{\tau_T^{k}}\right) \\& =\min_{T:T\#\mathbb{P}^T=\mathbb{P}_{\tau,\pi}}\mathbb{E}_{z\sim\mathbb{P}^T}\left[d_{x}\left(z,T\left(z\right)\right)\right]\\
 & =\frac{1}{N_T}\min_{T:T\#\mathbb{P}^T=\mathbb{P}_{\tau,\pi}}\sum_{n=1}^{N_T}d_{z}\left(z_{n},T\left(z_{n}\right)\right).
\end{align*}

Since $T\#\mathbb{P}^T=\mathbb{P}_{\tau,\pi}$, $T\left(z_{n}\right)=\tau_T^{k}$
for some $k$. Additionally, $\left|T^{-1}\left(\tau_T^{k}\right)\right|,k=1,...,K$
are proportional to $\pi_{k},k=1,...,K$. Denote $\sigma:\left\{ 1,...,N_T\right\} \rightarrow\left\{ 1,...,K\right\} $
such that $T\left(z_{n}\right)=\tau_T^{\sigma\left(n\right)},\forall i=1,...,N$,
we have $\sigma\in\Sigma_{\pi}$. It follows that 
\begin{align*}
    &\mathcal{W}_{d_{z}}\left(\frac{1}{N}\sum_{n=1}^{N_T}\delta_{z_{n}},\sum_{k=1}^{K}\pi_{k}\delta_{\tau_T^{k}}\right)\\&=\frac{1}{N_T}\min_{\sigma\in\Sigma_{\pi}}\sum_{n=1}^{N_T}d_{z}\left(z_{n},\tau_T^{\sigma\left(n\right)}\right).
\end{align*}

Finally, the the optimal solution of the OP in Eq. (\ref{eq:clustering_apd})
is equivalent to
\[
\min_{\mathcal{T,\pi}}\min_{\sigma\in\Sigma_{\pi}}\sum_{n=1}^{N_T}d_{z}\left(z_{n},\tau_T^{\sigma(n)}\right),
\]
which directly implies the conclusion.

\end{document}